\documentclass[conference]{IEEEtran}
\IEEEoverridecommandlockouts
\usepackage{cite}
\usepackage{amsmath,amssymb,amsfonts}
\usepackage{graphicx}
\usepackage{booktabs}
\usepackage{xcolor}
\usepackage{url}
\graphicspath{{figures/}}

\newcommand{\nframes}{167}     \newcommand{\nvideos}{9}
\newcommand{\nframesMain}{148} \newcommand{\nvideosMain}{8}
\newcommand{\maeSAM}{70.4}   \newcommand{\ciSAM}{56.0--86.1}   \newcommand{\rmseSAM}{118.2}  \newcommand{\recSAM}{0.54} \newcommand{\biasSAM}{-0.23}
\newcommand{\maeYOLO}{92.0}  \newcommand{\ciYOLO}{78.0--107.2} \newcommand{\rmseYOLO}{128.9} \newcommand{\recYOLO}{0.28} \newcommand{\biasYOLO}{-0.59}
\newcommand{\maeAPG}{152.9}  \newcommand{\ciAPG}{134.4--171.6} \newcommand{\rmseAPG}{192.0}  \newcommand{\recAPG}{2.14} \newcommand{\biasAPG}{+1.39}
\newcommand{\denseSAM}{308.1} \newcommand{\denseYOLO}{304.8} \newcommand{\denseAPG}{114.9}
\newcommand{\extSAM}{1{,}698.1} \newcommand{\extYOLO}{1{,}699.1} \newcommand{\extAPG}{391.7}
\newcommand{\crossoverDensity}{300} \newcommand{\oracleMAE}{44.1}
\newcommand{\corrSAM}{-0.89} \newcommand{\corrYOLO}{-0.64} \newcommand{\corrAPG}{0.52}

\begin{document}

\title{HAJJv2-CrowdCount\\
{\Large Zero-Shot Benchmark for Dense Crowd Counting}}

\author{
\IEEEauthorblockN{Reem AlYabis\IEEEauthorrefmark{1}, Fares AlTuwaim\IEEEauthorrefmark{1}, AlJawharh AlOtaibi\IEEEauthorrefmark{1}, Mohamed Eltahir\IEEEauthorrefmark{1}}
\IEEEauthorblockA{Reem@daldata.ai, Fares@daldata.ai, AlJawharh@daldata.ai, M.eltayeb@daldata.ai\\
Riyadh, Saudi Arabia}
}

\maketitle

\begin{abstract}
Automated crowd counting in Hajj video is difficult not because current models lack capacity, but because the footage violates the assumptions those models were built on: cameras observe the crowd from steep, near-vertical angles, individuals occlude one another extensively, and a single frame can contain well over a thousand people. Benchmarks that test crowd counting in such environment are either private or not detailed per second. We revisit the HAJJv2 dataset and contribute \textbf{HAJJv2-CrowdCount}: per-second human-annotated crowd counts for its testing videos\footnote{Per-second crowd-count annotations available at: \url{https://github.com/reem-8899/HAJJv2-CrowdCount}}. Using these annotations, we benchmark three recent zero-shot counting paradigms: an open-vocabulary detector (YOLO-World), a point-based counter (APGCC), and a promptable segmentation-based counter (SAM3Count). SAM3Count attains the lowest overall mean absolute error (MAE \maeSAM, 95\% CI \ciSAM), ahead of YOLO-World (\maeYOLO) and APGCC (\maeAPG). This ordering reverses, however, in the regime most relevant to deployment: on the densest frames, the detection- and segmentation-based counters both degrade sharply (MAE exceeding 300), while the point-based counter degrades far more gracefully (MAE \denseAPG). This inversion is decision-relevant for Hajj crowd management, where reliable counts are needed most precisely in the densest and most occluded scenes. The annotations are released to support reproduction and extension of these results.
\end{abstract}


\section{Introduction}
Crowd management during Hajj is fundamentally a safety problem. Accurate, near-real-time estimates of how many people occupy a corridor or courtyard directly inform operational decisions such as gate timing, flow direction, and crowd holds. Automated counting from existing camera infrastructure is therefore an attractive capability, but Hajj footage presents conditions under which most counting methods perform poorly: crowds are extremely dense, viewpoints are frequently near-vertical, and individuals are persistently occluded by one another.

A central practical question for a team seeking near-term deployment is whether a recent, general-purpose model can be applied directly, zero-shot without Hajj-specific training, and still achieve accuracy sufficient for an operations dashboard. Answering this question requires two resources that do not currently exist together: a Hajj test set with reliable per-frame counts, and a controlled, like-for-like comparison of current models evaluated on it.

This paper provides both. First, we annotate the HAJJv2 testing videos~\cite{hajjv2} with per-second crowd counts and release them publicly. Second, we benchmark three recent models on these annotations under an identical zero-shot protocol, reporting not only aggregate accuracy but where each model fails, which we show is precisely where an aggregate ranking becomes misleading. 

\section{Related Work}
Density-map regression has been the dominant paradigm in crowd counting since CSRNet~\cite{csrnet} demonstrated that dilated convolutions on a VGG backbone can produce sharp density maps for congested scenes. Subsequent work has refined this approach considerably. More recently, point- and attention-based formulations such as APGCC~\cite{apgcc} have improved accuracy on standard ShanghaiTech-style benchmarks by predicting head locations directly rather than only a scalar total.

In parallel, open-vocabulary detectors that accept a text prompt, e.g., YOLO-World~\cite{yoloworld}, enable person detection without training a dedicated detector, reducing counting to the enumeration of predicted boxes. This is operationally appealing because a single general-purpose model can be repurposed across tasks, but detection-by-box is known to degrade as scenes become denser and bounding boxes increasingly overlap.

The Segment Anything family, and counting methods built on it such as SAM3Count~\cite{sam3count}, constitute a third paradigm: the model is prompted for the object of interest and the resulting instance masks are counted.
 
HAJJv2~\cite{hajjv2} provides annotated Hajj crowd video and has previously been used for abnormal-behavior and flow analysis. To our knowledge, per-second total-count annotations suitable for a counting benchmark have not previously been released for its testing split. We address this gap.

\section{Per-Second Annotations for HAJJv2}
\label{sec:annotations}
\begin{figure}[t]
\centering
\includegraphics[width=\linewidth]{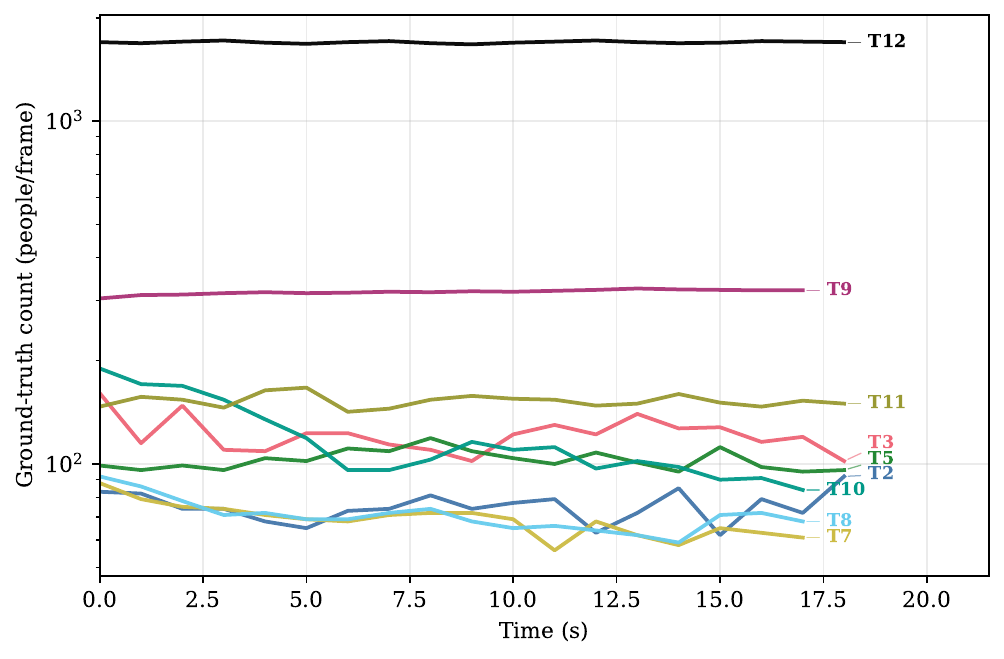}
\caption{Per-second ground-truth counts for the nine HAJJv2 testing videos (log scale). The benchmark spans two orders of magnitude in density. Testing\_12 (${\sim}1{,}700$ people/frame) provides an extreme-density stress test.}
\label{fig:gtprofiles}
\end{figure}

HAJJv2 comprises nine testing videos of Hajj crowds captured at varied densities and camera angles, including several steep, near-vertical viewpoints. We sampled each video at one frame per second and instructed annotators to record the number of visible people in each sampled frame, yielding \nframes{} labeled frames across the \nvideos{} videos. Fig.~\ref{fig:gtprofiles} shows the resulting density profiles: seven videos range from approximately 56 to 330 people per frame, while one (Testing\_12) reaches approximately $1{,}700$, providing the benchmark with a genuine extreme-density regime.

Two labeling protocols were used during annotation. Under \emph{independent counting}, the annotator enumerates every visible person in the frame using the image alone. Under \emph{model-assisted counting}, the annotator begins from a detector's output for the frame and adds any people the detector missed. Both protocols constitute legitimate annotation practice, and the latter is common in crowd labeling because it is more efficient on frames where the detector's output is already close to correct. Of the nine videos, six were labeled independently and three (Testing\_2, Testing\_3, Testing\_5) were labeled with model assistance.

Only individuals whose appearance was sufficiently clear within the frame were included in the count. Persons who were partially visible, heavily occluded, or reduced to indistinguishable distant objects by perspective distortion were excluded until they became clearly visible in a subsequent frame. Each frame was inspected systematically to mark every counted individual exactly once: a person entering the scene was counted only once fully visible, and a person leaving the scene was removed from the count once no longer fully visible. In regions where individuals temporarily occluded one another, only clearly distinguishable individuals were counted, with fully occluded persons excluded until they reappeared. Perspective distortion, varying camera viewpoints, and dense occlusion required particular care to distinguish valid individuals from ambiguous visual artifacts and to apply these rules consistently across the dataset; the same visibility and occlusion criteria were applied when annotators added detector-missed individuals under the model-assisted protocol.
 
To improve annotation reliability, every frame was counted twice by the same annotator. Whenever the two passes disagreed, the frame was re-examined and a verified count was recorded. This double-pass procedure reduced counting inconsistencies and served as an additional quality-control step prior to finalizing the annotations.

\section{Models Compared}
\begin{figure*}[t]
\centering
\begin{minipage}{0.32\textwidth}\centering
\includegraphics[width=\linewidth]{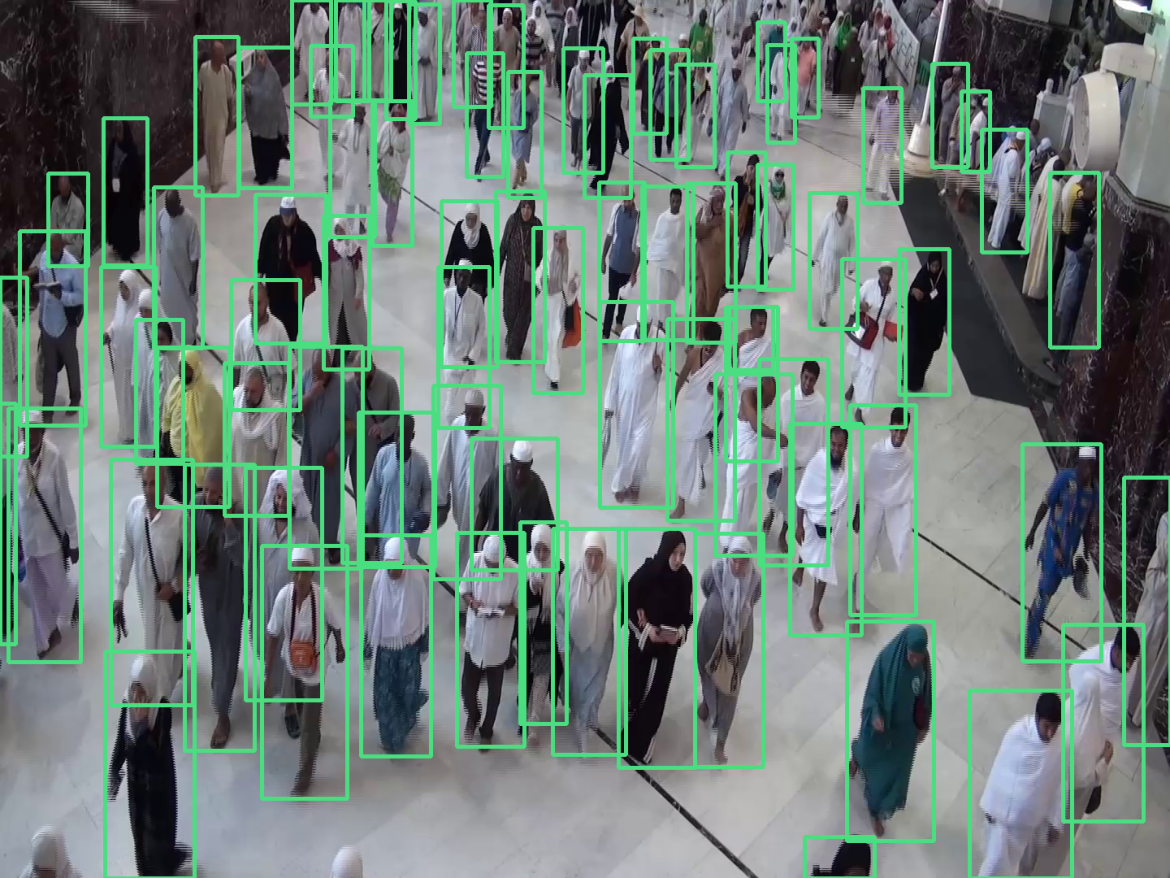}\\{\footnotesize (a) YOLO-World - bounding boxes}
\end{minipage}\hfill
\begin{minipage}{0.32\textwidth}\centering
\includegraphics[width=\linewidth]{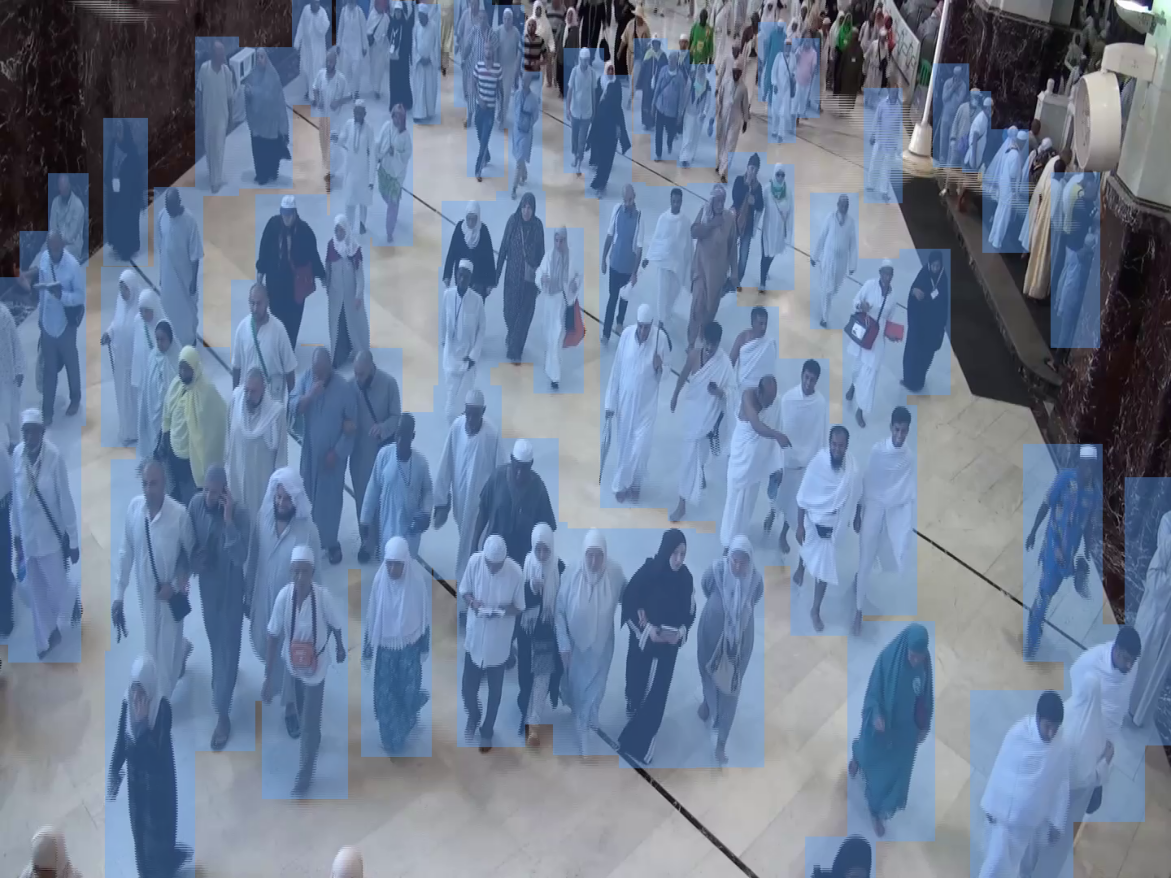}\\{\footnotesize (b) SAM3Count - segmentation masks}
\end{minipage}\hfill
\begin{minipage}{0.32\textwidth}\centering
\includegraphics[width=\linewidth]{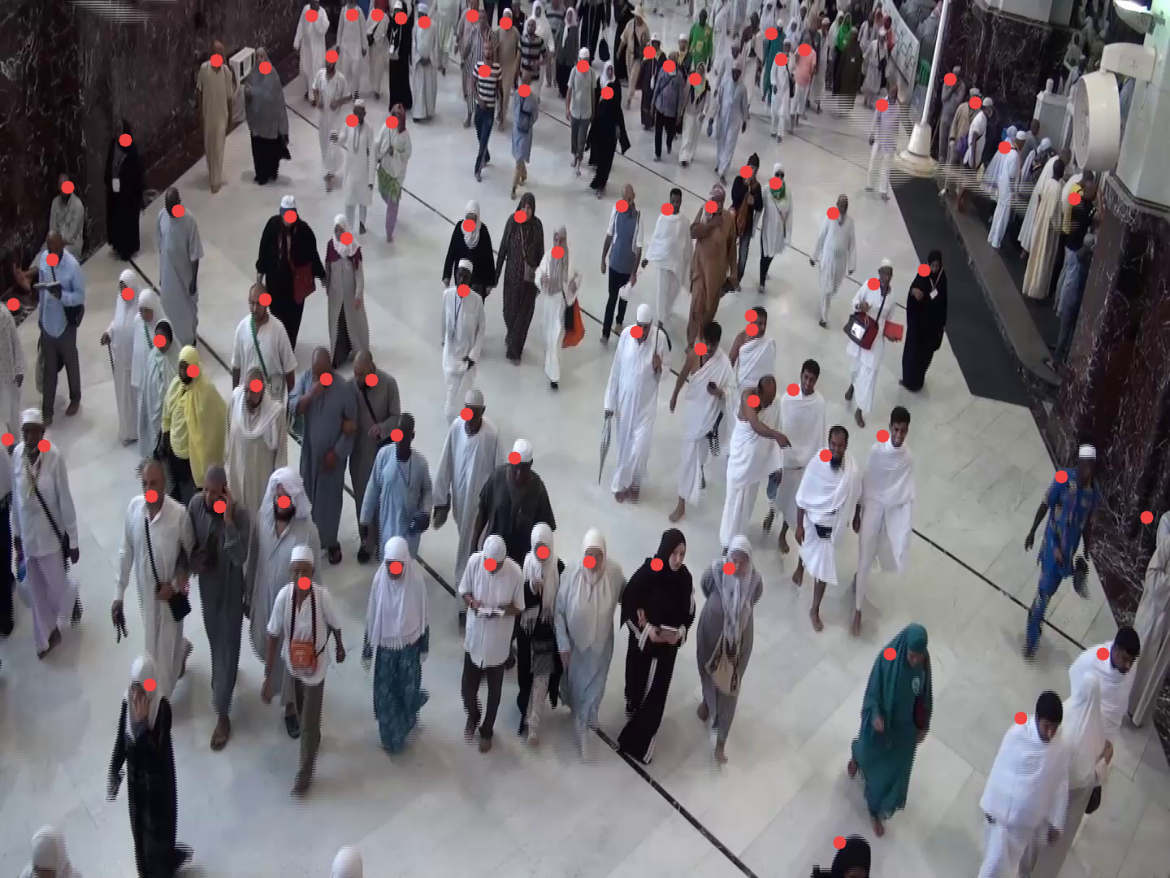}\\{\footnotesize (c) APGCC - point centroids}
\end{minipage}
\caption{The three counting modalities on a dense HAJJv2 frame. (a) YOLO-World detects boxes reliably in the foreground but misses small background figures below its resolution limit. (b) SAM3Count delineates irregular pilgrim boundaries, with mask merging in the most congested regions. (c) APGCC predicts head centroids, the modality best matched to top-down occlusion. As the density-band analysis in Sec.~\ref{sec:bands} shows, it is also the most robust of the three on the densest frames, despite a global tendency to over-count.}
\label{fig:qual}
\end{figure*}

We benchmark one representative of each of three current counting paradigms. Fig.~\ref{fig:qual} shows the three output modalities on the same dense HAJJv2 frame. All are evaluated strictly zero-shot (Sec.~\ref{sec:protocol}).

\subsection{YOLO-World (open-vocabulary detection)}
\label{sec:yolo}
YOLO-World~\cite{yoloworld} represents the open-vocabulary detection paradigm and serves as the operational baseline. It couples a YOLOv8 detection backbone with a CLIP text encoder through a re-parameterizable vision-language path (RepVL-PAN), so the model can be prompted for \texttt{person} boxes without person-specific training. We evaluate the \texttt{yolov8s-worldv2} checkpoint with confidence $0.10$ and IoU $0.45$, selected on a single held-out calibration video (the default confidence of $0.25$ discards a substantial fraction of true detections in dense frames).

\begin{figure}[t]
\centering
\includegraphics[width=0.95\linewidth]{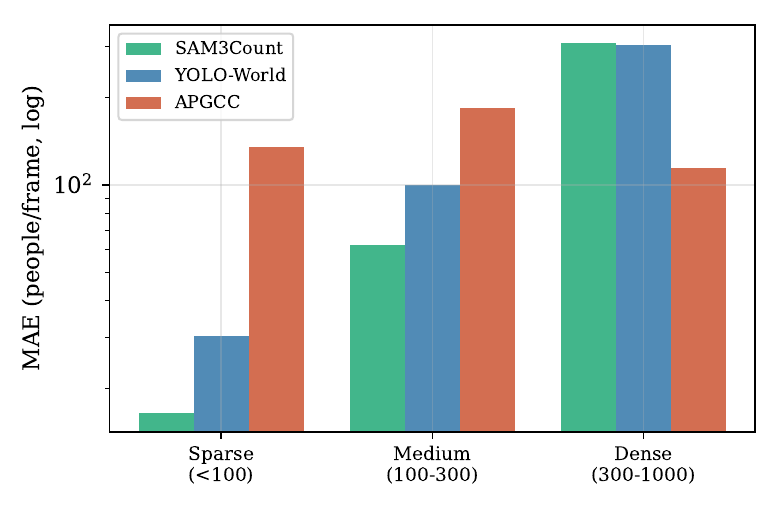}
\caption{MAE by density band (log scale), shown here for all three models for comparison. SAM3Count and YOLO-World are strongest on sparse and medium frames but both degrade sharply in the dense band. APGCC (introduced in Sec.~\ref{sec:apgcc}) over-counts on easier frames yet is by far the most accurate on dense frames.}
\label{fig:bands}
\end{figure}

\begin{figure}[t]
\centering
\includegraphics[width=0.8\linewidth]{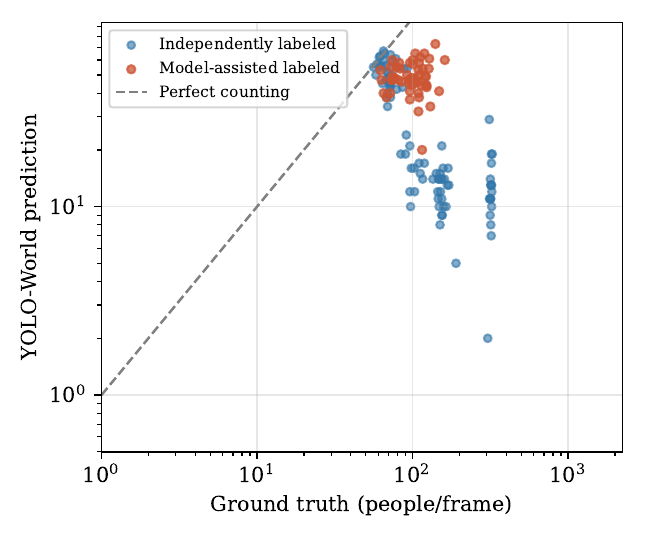}
\caption{YOLO-World: per-second prediction vs.\ ground truth (log-log), colored by annotation protocol. The dashed line is perfect counting. Predictions flatten as density grows.}
\label{fig:yoloscatter}
\end{figure}

Because counting reduces to enumerating detected boxes, the paradigm is inherently sensitive to occlusion: box detection presumes a largely visible body silhouette. Fig.~\ref{fig:yoloscatter} plots every per-second prediction against ground truth. Predictions increasingly deviate from the identity line as density increases, and Fig.~\ref{fig:bands} quantifies this degradation by density band, with MAE rising from $30$ on sparse frames to $100$ on medium frames and exceeding $300$ on dense frames, where perspective distortion additionally reduces far-field pilgrims below the detector's spatial resolution. The model under-counts throughout (bias \biasYOLO, recovery \recYOLO), consistent with boxes failing to fire on occluded bodies.

\subsection{SAM3Count (promptable segmentation counting)}
SAM3Count~\cite{sam3count} represents the segmentation paradigm: prompt a Segment Anything model with the concept \texttt{person} and count the resulting instance masks (we retain masks with confidence above $0.5$).
 
\begin{figure}[t]
\centering
\includegraphics[width=0.8\linewidth]{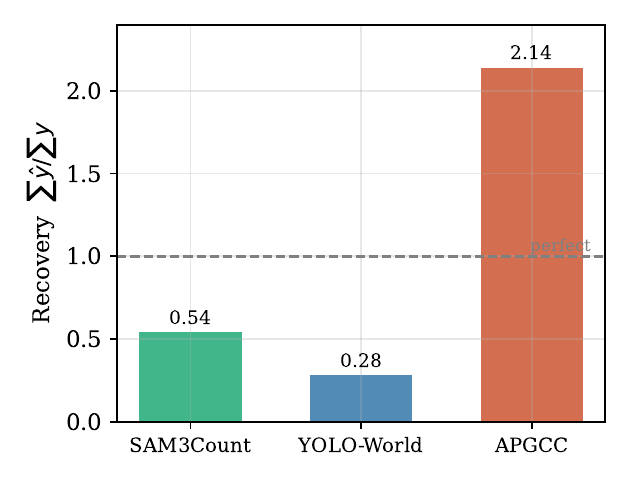}
\caption{Fraction of the total true count recovered by each model ($\sum\hat{y}/\sum y$. Dashed line $=$ perfect). SAM3Count and YOLO-World under-count (recovery \recSAM{} and \recYOLO). APGCC over-counts by roughly $2\times$ (recovery \recAPG).}
\label{fig:recovery}
\end{figure}

A Hajj-specific challenge for this paradigm arises because pilgrims wear the near-identical white ihram garment, which removes much of the color and texture contrast that ordinarily separates individuals in crowd footage. Despite this, masks prove considerably more robust than bounding boxes at low and medium density, since a partial mask can still register where a box fails to fire. SAM3Count recovers a larger fraction of the true count than the detector overall (recovery \recSAM{} versus \recYOLO, Fig.~\ref{fig:recovery}), yielding the lowest overall MAE in the comparison (Sec.~\ref{sec:results}). Its principal failure mode is \emph{mask merging}, in which adjacent pilgrims in the most congested regions fuse into a single mask, producing localized under-counts that increase with density. As Sec.~\ref{sec:bands} shows, this effect is severe enough that SAM3Count loses its advantage precisely on the densest frames.

\subsection{APGCC (point-based counting)}
\label{sec:apgcc}
APGCC~\cite{apgcc} represents the point-based paradigm, localizing head centroids directly as two-dimensional points under auxiliary anchor-point guidance rather than regressing boxes, masks, or density maps. This modality is, in principle, well matched to Hajj footage: in a steep, top-down view of a packed crowd, the head is often the only consistently visible body part. We evaluate the standard ShanghaiTech-pretrained weights zero-shot.

The zero-shot transfer is imperfect but instructive. Overall, APGCC has the highest MAE of the three models (\maeAPG) and over-counts by approximately a factor of two (bias \biasAPG, recovery \recAPG), predicting substantially more heads than are present on sparse and medium-density frames. This global over-count, however, is also the source of its robustness where the other two paradigms fail: on the densest frames, the box- and mask-based counters lose most of the crowd to occlusion and mask merging, whereas APGCC's head-point predictions continue to track the true count. Its dense-band MAE is \denseAPG, compared with \denseSAM{} for SAM3Count and \denseYOLO{} for YOLO-World (Sec.~\ref{sec:bands}), and on the extreme-density video its error is smaller than either competitor's by a factor of four (Sec.~\ref{sec:extreme}), making it more suitable for very dense scenes.

\section{Evaluation Protocol}
\label{sec:protocol}
For each model, we produce one predicted count per labeled frame and compare it to the corresponding human count. We report mean absolute error (MAE) and root mean squared error (RMSE) in people per frame, together with recovery, defined as the fraction of the total true count recovered by the model ($\sum\hat{y}/\sum y$), and mean signed relative bias. A 95\% confidence interval for MAE is obtained by bootstrapping over frames ($2{,}000$ resamples). Because operational risk increases with crowd density, we additionally report MAE within three density bands, sparse ($<100$), medium ($100$--$300$), and dense ($300$--$1{,}000$ people per frame), so that a model accurate only on sparse frames is not favored by an aggregate average. Because the extreme-density video Testing\_12 (${\sim}1{,}700$ people per frame) dominates any pooled average on its own, the headline results in Table~\ref{tab:main} cover the remaining \nvideosMain{} videos (\nframesMain{} frames); Testing\_12 is reported separately in Sec.~\ref{sec:extreme}.

\section{Results}
\label{sec:results}
 
\begin{table}[t]
\centering
\caption{Zero-shot crowd counting on HAJJv2 (\nvideosMain{} testing videos, \nframesMain{} frames. The extreme-density video is reported separately in Sec.~\ref{sec:extreme}). Lower MAE/RMSE is better. Recovery and bias near $1.0$ and $0$ are better.}
\label{tab:main}
\begin{tabular}{lcccc}
\toprule
Model & MAE (95\% CI) & RMSE & Recovery & Bias \\
\midrule
\textbf{SAM3Count} & \textbf{\maeSAM} (\ciSAM) & \rmseSAM & \recSAM & \biasSAM \\
YOLO-World & \maeYOLO{} (\ciYOLO) & \rmseYOLO & \recYOLO & \biasYOLO \\
APGCC & \maeAPG{} (\ciAPG) & \rmseAPG & \recAPG & \biasAPG \\
\bottomrule
\end{tabular}
\end{table}

\begin{figure}[t]
\centering
\includegraphics[width=0.8\linewidth]{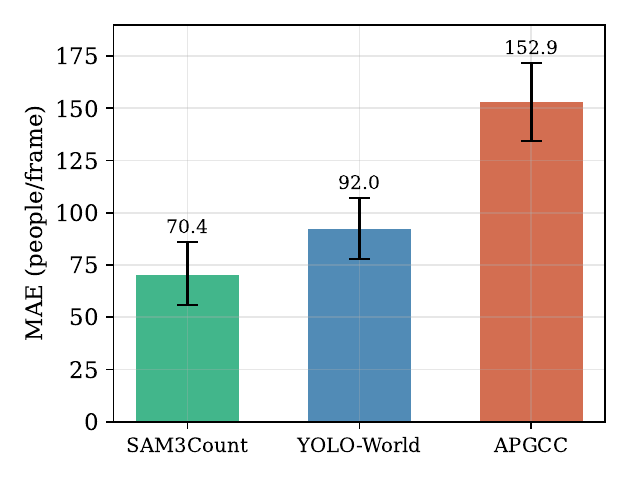}
\caption{Overall MAE with bootstrap 95\% confidence intervals (main set, \nframesMain{} frames).}
\label{fig:maemodels}
\end{figure}

Table~\ref{tab:main} and Fig.~\ref{fig:maemodels} report the metrics. SAM3Count obtains the lowest overall MAE (\maeSAM, 95\% CI \ciSAM), ahead of YOLO-World (\maeYOLO) and APGCC (\maeAPG). The gap between SAM3Count and YOLO-World is modest. The recovery and bias columns account for this ordering: SAM3Count and YOLO-World both under-count overall (recovery \recSAM{} and \recYOLO), with segmentation masks registering more partially occluded pilgrims than bounding boxes do, while APGCC over-counts by approximately $2\times$ (recovery \recAPG).

\subsection{Density bands}
\label{sec:bands}
\begin{figure}[t]
\centering
\includegraphics[width=\linewidth]{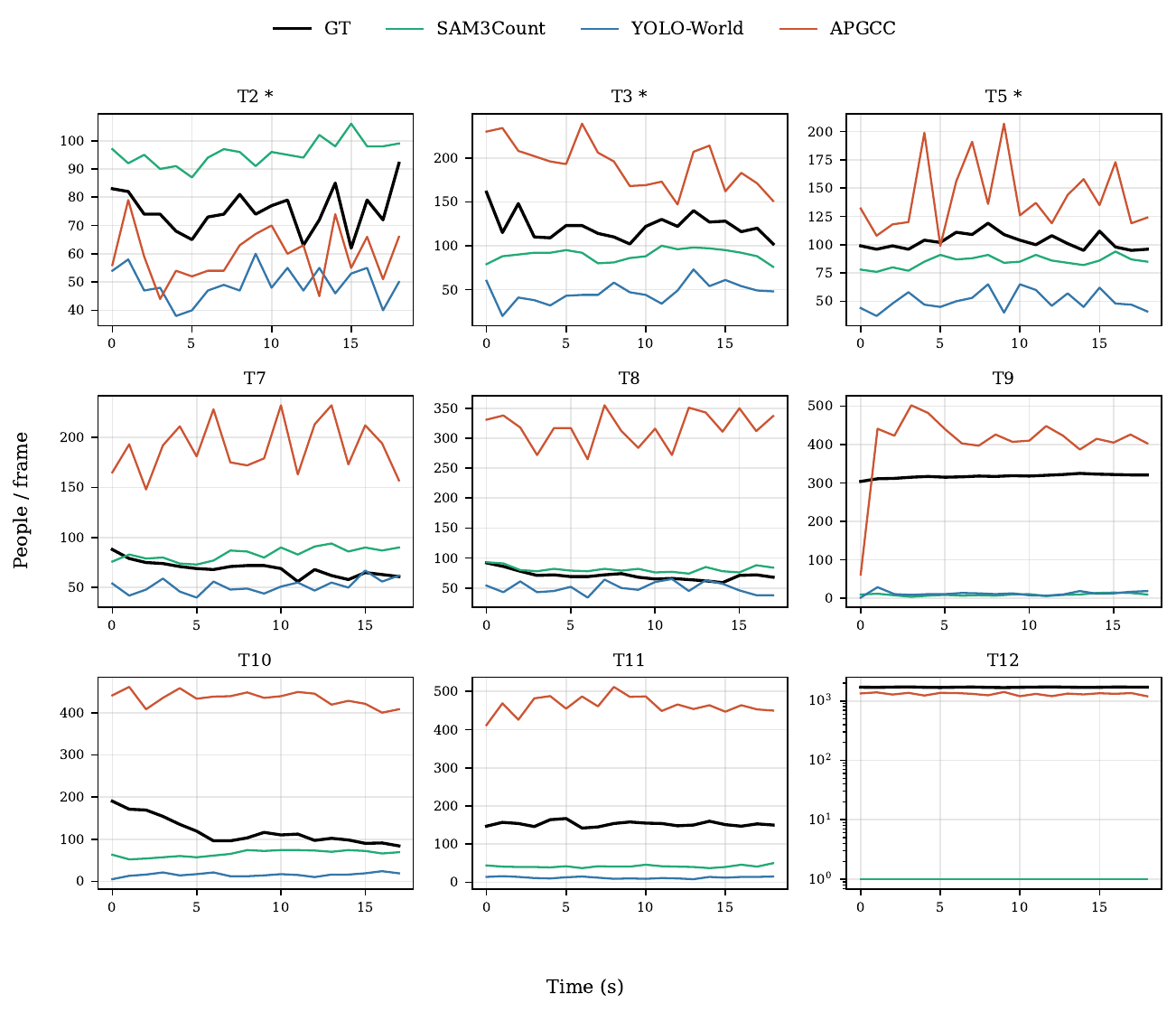}
\caption{Per-second traces for all nine testing videos (asterisked titles are model-assisted): ground truth (black), SAM3Count (green), YOLO-World (blue), and APGCC (red). On the four densest videos (Testing\_8 through Testing\_11) and on the extreme video Testing\_12, SAM3Count and YOLO-World both collapse toward zero while APGCC over-shoots but continues to track the crowd's shape.}
\label{fig:timeseries}
\end{figure}

The pattern across density bands is the most decision-relevant finding in this study, and it reverses the overall ranking. On sparse and medium frames, the ordering matches the aggregate result: SAM3Count is most accurate (MAE $16.4$ and $62.1$), followed by YOLO-World ($30.3$ and $100.2$), with APGCC least accurate ($135.8$ and $184.3$), reflecting its tendency to over-count on easier scenes. On dense frames ($300$-$1{,}000$ people), however, the ordering reverses: the box- and mask-based counters converge to a similarly large error (YOLO-World \denseYOLO, SAM3Count \denseSAM) as boxes fail to fire and masks merge under heavy occlusion, whereas APGCC's head-point predictions remain substantially more accurate, with an MAE of \denseAPG{}, less than half that of either competitor (Fig.~\ref{fig:bands}). Fig.~\ref{fig:timeseries} shows the corresponding per-video traces. This reversal is operationally significant: a Hajj corridor under strain is, by definition, a dense and heavily occluded scene, and it is precisely in this regime that the paradigm with the weakest overall average proves the most reliable.

\subsection{Where the ranking flips, and what a perfect router would gain}
\label{sec:crossover}
\begin{figure}[t]
\centering
\includegraphics[width=\linewidth]{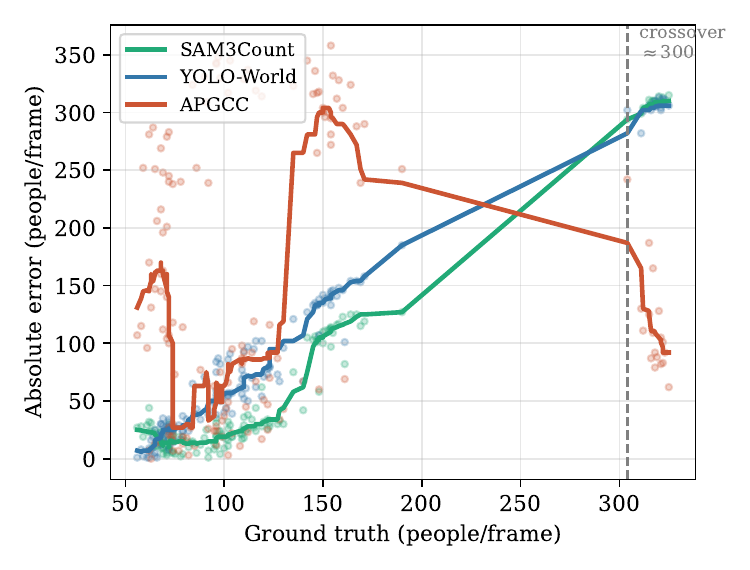}
\caption{Absolute error versus ground-truth density (excluding Testing\_12), with per-frame values (light) and a rolling-median trend (bold) for each model. SAM3Count's and YOLO-World's error grows roughly linearly with density. APGCC's error is non-monotonic but falls below both competitors beyond the crossover at approximately \crossoverDensity{} people per frame.}
\label{fig:crossover}
\end{figure}

The three density bands in Sec.~\ref{sec:bands} establish that the ranking reverses somewhere between the medium and dense regimes, but a bin boundary is a coarse instrument for locating the reversal itself. Fig.~\ref{fig:crossover} plots absolute error against ground-truth density continuously (excluding Testing\_12) with a rolling median trend per model. The crossover is sharp rather than gradual: SAM3Count and YOLO-World's error rises roughly linearly with density throughout, while APGCC's error is non-monotonic, spiking near $130$--$170$ people per frame before falling as density increases further. The two trends intersect at approximately \crossoverDensity{} people per frame, giving a concrete answer to a question the banded analysis could only bracket.
 
This reversal raises a natural question: how much accuracy is left on the table by committing to a single model rather than selecting the best one per frame? An oracle that selects, for each frame, whichever of the three models is closest to the true count achieves an MAE of \oracleMAE, a $37\%$ reduction relative to SAM3Count alone (\maeSAM). This gap is a ceiling, not a demonstrated method, but it indicates that density-aware model selection is a meaningfully open direction rather than a marginal one.
 
Building such a selector requires a signal that predicts which regime a frame is in without access to the ground truth it is trying to estimate. A natural candidate is a model's own raw output, and here the paradigms diverge in an unexpected way. Table~\ref{tab:corr} reports the Pearson correlation between each model's raw prediction and the true count. SAM3Count's raw count is \emph{negatively} correlated with true density ($r=\corrSAM$): as crowds become denser, progressive mask merging reduces the number of distinct instances detected, so the model's own output moves opposite to the quantity it is meant to track. YOLO-World shows the same inversion, driven by the same occlusion mechanism acting on boxes ($r=\corrYOLO$). Only APGCC's raw output remains positively associated with true density ($r=\corrAPG$), despite its systematic over-count. Consequently, neither SAM3Count nor YOLO-World can supply its own confidence or count as a proxy for ``how dense is this scene,'' which rules out the most obvious self-referential router and points toward an external density estimator as a prerequisite for closing the gap toward the oracle bound.
 
\begin{table}[t]
\centering
\caption{Pearson correlation between each model's raw predicted count and the true count (excl. Testing\_12). A negative value indicates the model's own output moves opposite to true crowd size as density increases.}
\label{tab:corr}
\begin{tabular}{lc}
\toprule
Model & $r$(prediction, ground truth) \\
\midrule
SAM3Count & \corrSAM \\
YOLO-World & \corrYOLO \\
APGCC & \corrAPG \\
\bottomrule
\end{tabular}
\end{table}

\subsection{The extreme regime}
\label{sec:extreme}
On Testing\_12 (${\sim}1{,}700$ people per frame), the mask- and box-based paradigms are not merely inaccurate but uninformative: SAM3Count's mask merging drives its error to MAE ${\approx}\extSAM{}$, and YOLO-World detects almost no one (MAE ${\approx}\extYOLO{}$). In both cases, the error is comparable to the size of the crowd itself. APGCC, despite over-counting elsewhere, is substantially closer here (MAE ${\approx}\extAPG{}$, roughly one-quarter of either competitor's error), although it still under-counts the true value of approximately $1{,}700$. This extreme-density regime, characteristic of peak-flow conditions at the Mataf, remains an open problem, and no model's zero-shot per-frame count should currently be trusted at this density. Nonetheless, the ordering observed in the dense band persists here as well, again favoring the point-based paradigm.

\section{Conclusion}
We have released HAJJv2-CrowdCount, per-second crowd-count annotations for the HAJJv2 testing videos, and used them to benchmark three recent models under a strictly zero-shot protocol. SAM3Count achieved the lowest overall MAE (\maeSAM), but the density-band analysis yields the more consequential finding: the model with the best overall average is not the model most accurate on the dense, occluded frames that a Hajj safety system depends on, where the point-based counter (APGCC) proved the most robust. No model evaluated here is deployment-ready in the extreme-density regime. The annotations are released publicly to support reproduction and extension of these results to additional models.

\end{document}